\documentclass[lettersize,journal]{IEEEtran}
\usepackage{amsmath,amsfonts}
\usepackage{algorithmic}
\usepackage{algorithm}
\usepackage{array}
\usepackage[caption=false,font=normalsize,labelfont=sf,textfont=sf]{subfig}
\usepackage{textcomp}
\usepackage{stfloats}
\usepackage{url}
\usepackage{verbatim}
\usepackage{graphicx}
\usepackage{cite}
\usepackage{mathrsfs}
\usepackage{amssymb}
\usepackage{footnote}
\usepackage{multirow}
\usepackage{tabularx}
\usepackage{booktabs}
\usepackage{utfsym}
\usepackage{bm}
\usepackage{mathrsfs}
\begin{document}

\title{Superpixel Semantics Representation and Pre-training for Vision-Language Tasks}

\author{Siyu~Zhang,
        Yeming~Chen,
        Yaoru~Sun$^{\ast}$,
        Fang~Wang,
        Jun~Yang,\\
        Lizhi~Bai,
        and Shangce~Gao
\thanks{Siyu Zhang, Yeming Chen, Yaoru Sun, Jun Yang, and Lizhi Bai are with the Department of Computer Science and Technology, Tongji University, Shanghai 201804, China (e-mail:\{zsyzsy, 2130769, yaoru, junyang, bailizhi\}@tongji.edu.cn).}
\thanks{Fang Wang is with Department of Computer Science, Brunel University, Uxbridge UB8 3PH, UK (e-mail:fang.wang@brunel.ac.uk).}
\thanks{Shangce Gao is with the Faculty of Engineering, University of Toyama, Toyama-shi 930-8555, Japan (e-mail:gaosc@eng.u-toyama.ac.jp).}
\thanks{$\ast$ Corresponding author.}}
\markboth{}%
{Shell \MakeLowercase{\textit{et al.}}: A Sample Article Using IEEEtran.cls for IEEE Journals}


\maketitle

\begin{abstract}
The key to integrating visual language tasks is to establish a good alignment strategy. Recently, visual semantic representation has achieved fine-grained visual understanding by dividing grids or image patches. However, the coarse-grained semantic interactions in image space should not be ignored, which hinders the extraction of complex contextual semantic relations at the scene boundaries. This paper proposes superpixels as comprehensive and robust visual primitives, which mine coarse-grained semantic interactions by clustering perceptually similar pixels, speeding up the subsequent processing of primitives. To capture superpixel-level semantic features, we propose a Multiscale Difference Graph Convolutional Network (MDGCN). It allows parsing the entire image as a fine-to-coarse visual hierarchy. To reason actual semantic relations, we reduce potential noise interference by aggregating difference information between adjacent graph nodes. Finally, we propose a multi-level fusion rule in a bottom-up manner to avoid understanding deviation by mining complementary spatial information at different levels. Experiments show that the proposed method can effectively promote the learning of multiple downstream tasks. Encouragingly, our method outperforms previous methods on all metrics. Our code will be released upon publication. 
\end{abstract}


\section{Introduction}
\IEEEPARstart{M}{ultimodal} tasks aim to enable computers with the ability to effectively and efficiently process multi-channel (visual and textual) signals, in order to learn the key concepts needed to better comprehend the world. Vision and language (VL), as important components of human perception, are currently popular research topics. Benefiting from the great success of self-supervised learning, vision and language pre-training (VLP) has become an increasingly central training paradigm in multimodal tasks. In recent years, VLP models have achieved state-of-the-art results in many VL downstream tasks, involving visual question answering (VQA) \cite{b1}, visual reasoning (VR) \cite{b2}, visual entailment (VE) \cite{b3}, and image-text retrieval \cite{b4}, etc.

Efficient visual representation helps close the semantic gap between visual language modalities. Earlier VL methods heavily rely on a set of bounding boxes, most of which involve region supervision (e.g., object detection). However, these methods usually represent a portion of the image while ignoring the coherence of semantic features, such as the spatial relations between objects with overlap, shapes of objects, and scenes. Recent methods have achieved successful scaling on VL tasks. These works revisit grid features \cite{b5}, \cite{b6}, \cite{b7} or divide image patches \cite{b8} to represent fine-grained visual information, and take an end-to-end manner to directly align the two modalities. Although they achieve better performance in overcoming the limitations of object detection. However, there is still a semantic gap in terms of visual understanding. On the one hand, pixels are vulnerable to noise and distortion, which complicates the task of extracting homogeneous regions from an image. On the other hand, image features at the pixel-level are more diverse and detailed than highly abstract and concise textual tokens. As a result, aligning VL information becomes a challenging task. 
\IEEEpubidadjcol

To resolve the above issues, we redefine a visual semantic optimizer, which can be flexibly plugged into any multimodal model. Our optimizer consists of three important level hierarchies, each covering visual semantic features of different regions. Specifically, the input raw image is first transformed into superpixel space. Superpixels are homogeneous regions composed of multiple contiguous pixels, which can protect structure cues with low complexity. Secondly, a Multiscale Difference Graph Convolutional Network (MDGCN) is adopted to model superpixel-level spatial topological features in a region-merging manner. In this way, it not only understands the contextual semantic relations of objects but also improves the scalability problem on the graph. Notably, we focus on the subtle differences between adjacent objects, which provide valuable information for aggregating graph node features. Then, we use CNNs to learn continuous latent variables at the pixel-level. It alleviates the problem that graph nodes ignore local spatial information at individual pixels due to the property of superpixel-level features of GCNs to identify homogeneous regions. Finally, we design a multi-level parsing architecture (pixel and superpixel-level) with a bottom-up fusion strategy to build visual representations. The aligned vision and language features can be fed into VLP models for better application in downstream tasks.

Different from previous works, we are devoted to using superpixel as the visual primitives for traversability instead of using patches. Superpixel can naturally perceive object contours (e.g., occlusion edges) while helping to minimize the data required for representing images. In particular, superpixel exploits the spatial continuity of image features to reduce noise, which is more coherent than the patch. Furthermore, we construct a robust MDGCN model. For one thing, the difference operation can filter out potential noise caused by irregular primitive edges. The model employs central difference graph convolution operation to predict differences between the features of the central node and the adjacent nodes. For another, the analysis of multiscale features at the superpixel-level facilitates a comprehensive understanding of the spatial topology with limited training samples and the subtle hierarchical relations among scattered and complex objects. In pursuit of comprehensive image understanding to facilitate multimodal alignment, we enrich semantic features from global and local multi-level perspectives and integrate them into the encoder. We offer more details on implementation and conduct extensive experiments to validate the model performance. Experimental results show that our method achieves more solid performance than other state-of-the-art models. 

In a nutshell, the main contributions of this work can be concluded as follows:

\begin{itemize}[\IEEEsetlabelwidth{Z}]
\item We propose using superpixel as the visual primitives to achieve coarse-grained semantic interactions, which are more suitable than patch regions for capturing semantic information in complex scenes. To the best of our knowledge, we are the first to reformulate the VL-based alignment task as a superpixel-level VL pre-training task.
\item We construct a multiscale difference graph convolutional network, termed MDGCN, which aggregates local gradient difference information between adjacent graph nodes by computing central differential convolution operators to further improve object semantic contrast.
\item To strengthen visual interpretations of image structure, we provide a learning strategy that integrates GCN and CNN multi-level visual representations, aiming at learning superpixel-level and pixel-level semantic features composition, relational, and hierarchical structures of visual inter-modality.
\item Experiments prove the effectiveness of our method on multiple VL downstream tasks. The proposed scheme outperforms state-of-the-art models on all metrics.\\
\end{itemize}

The remainder of this paper is organized as follows. In Section II, we introduce the related works. In Section III, the proposed MDGCN method with details is elaborated. The extensive experiments and ablation studies are exhibited and analyzed in Section IV. Finally, the conclusion is presented in Section V.

\section{Related Work}
\subsection{Visual Representation Schema} 
With the great success of deep learning in the computer vision (CV) and natural language processing (NLP) \cite{b9} fields. People are no longer satisfied with the processing of a single modality. Instead, multimodal comprehension tasks get much attention. Vision is a large part of how humans perceive the environment, while language-aligned visual features can effectively improve the performance of VL. Unlike the discrete signal spaces (e.g., words or sub-word units) built in tokenized vocabularies for languages, image raw signals are typically continuous and high-dimensional. Consequently, image tokenization is a more complex process than textual tokenization. Reviewing the development of VL models, image tokenization can generally be divided into three stages: region-based, grid-based, and patch-based.

\emph{Region Feature}. Early visual feature detection was obtained from a pre-trained object detector. Shih \textit{et al.}\cite{b10} first focused on the idea of region features and applied it to VQA. Anderson \textit{et al.}\cite{b11} developed a combined bottom-up and top-down attention (BUTD) model, which used Faster R-CNN pre-trained on the Visual Genome (VG) \cite{b12} dataset as region feature embedding. After that, region-based features quickly dominated the design of most VL tasks. The advantage of regional features lies in the simplicity of object location representation and more focus on salient image regions. However, the visual features are constrained to specific region categories by object detection methods, thereby limiting the scalability and generalization capabilities of VLP models. Furthermore, some important factors of visual information are lost, making it difficult to understand broader semantic relations.

\emph{Grid Feature}. Recently, some works have revisited grid-based features, which perform convolution operations on equally sized image grids, as effective visual representations. For example, Jiang \textit{et al.}\cite{b5} proposed a grid-based VQA model that can achieve equally performant region features. Cho \textit{et al.}\cite{b13} leveraged grid features as the visual inputs to enable LXMERT to generate images. To learn effective visual information directly from pixel features, Huang \textit{et al.}\cite{b6} attempted to use Pixel-BERT to break this limitation. Instead of extracting all pixels for pre-training, they only randomly sample a portion of them. However, random sampling only slightly improves performance by less than 0.5 in the VQA task. After that, Huang \textit{et al.}\cite{b7} leveraged a CNN-based grid features end-to-end pre-training framework that learned concrete yet compact visual representations based on a vision dictionary. However, the adjacent grids are mapped to the same vision dictionary by their designed MVM, which causes embedded vectors to be lost during the masking. Although grid features eliminate the need for bounding boxes and are convenient to use, deeper networks come with a higher computational burden.

\emph{Patch Feature}. Patch features are typically obtained through linear projections applied to pixel patches of a fixed size. The concept of patch-based was first proposed by Vision Transformer (ViT) \cite{b14}, which motivated researchers to present ViT-based VLP models. For example, textual tokens and image patches are directly employed by \cite{b8} for the pre-trained ViT model. In addition, METER\cite{b15}, X-VLM\cite{b16}, ALBEF\cite{b17}, and BLIP\cite{b18} all adopt ViT as their visual encoder. In contrast to grid features, patch features can be simply embedded using a linear projection, which improves running efficiency. However, patch features are susceptible to noise, which reduces classification accuracy. Besides, fixed-sized patches can span multiple different visual regions, which makes it difficult to learn subtle features of complex object boundaries. 
\begin{figure*}[!t]
\centering
\includegraphics[width=6.8in]{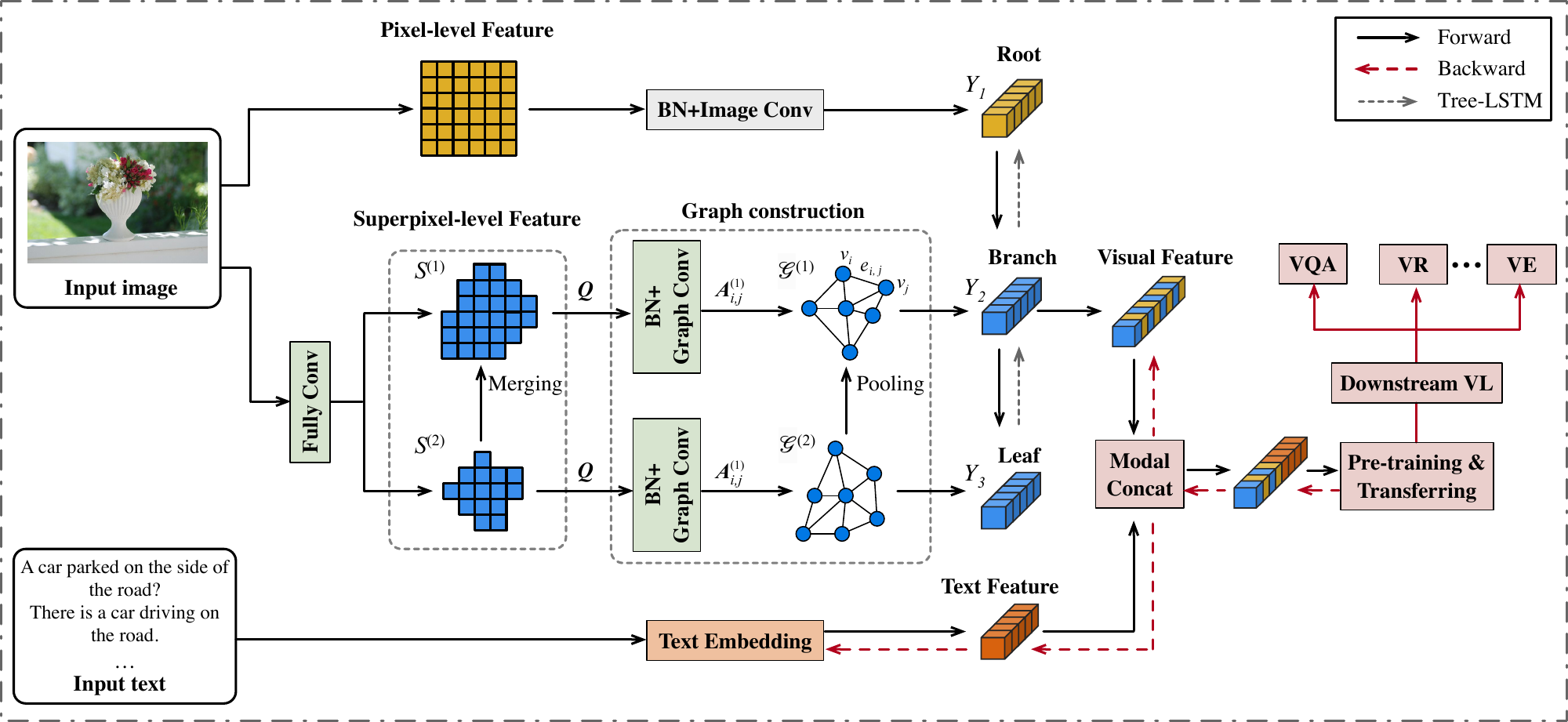}
\caption{The framework of our scheme to integrate pixel- and superpixel-based complementary information for the VL alignment task. Note that BN denotes batch normalized input features, and “modal concat” means multi-modal concatenation.}
\label{fig_1}
\end{figure*}
\subsection{Visual and Language Tasks} 
There exists a large amount of research work on multi-modal VL tasks. Early VL methods mainly adopted a simple feature fusion, such as element-wise addition, or product operations. Later on, most attention-based methods were developed to find more key clues by assigning different weights to detection objects. For example, Peng \textit{et al.} \cite{b19} designed a composed relation attention network, which can describe more sophisticated trinary relations. In \cite{b20}, a re-attention model was developed that efficiently assisted with visual attention learning by minimizing consistency loss. While models used for VL have shown improvement in performance, they still face challenges, especially in counting and reasoning.

With further development, some works attempt to improve VL model performance with graph-structured representations. In the work of \cite{b21}, an answer-centric Radial-GCN approach localized core answer regions by matching semantic labels detected in an image with latent answers. Chen \textit{et al.}\cite{b22} proposed graph convolutional autoencoders to learn rich visual features, which can encode the relationship between pedestrians and their surrounding objects to reason crossing intention. Guo \textit{et al.}\cite{b23} proposed bilinear graph networks to learn the context of the joint embedding of objects and text, enabling multistep reasoning. Recently, most researchers have worked on transformer-based single-stream models, which can perform inter-modality and intra-modality fusion. 

Chen \textit{et al.}\cite{b24} provided a single-stream VLP model named UNITER, which adopted the simple yet powerful Transformer model as the backbone. Similar to UNITER, several works such as OSCAR \cite{b25} and VILLA \cite{b26} also used single-stream modeling. Kim \textit{et al.}\cite{b8} presented the vision-and-language transformer (ViLT), a minimal VLP model designed for multimodal downstream tasks. In \cite{b27}, an mPLUG architecture was provided, which introduced an asymmetric structure to align VL features and improve computation efficiency. Notwithstanding the impressive performance achieved by these methods, the formidable challenges posed by the large dataset size and high computational cost hinder the training of the model.
 
Despite the above work has developed a variety of multimodal alignment methods, this is the first time that superpixel-level and pixel-level features are combined in one structure for visual semantic alignment. In addition, the MDGCN-based model is applied to analyze the structure and geometry information, which can reflect the differences among graph nodes in the neighborhood. 

\section{Methodology}
In this section, we elaborate on the proposed method. More intuitively, the overview schematic framework is shown in Fig.~\ref{fig_1}. The specific process mainly consists of four stages, including superpixel-level representation, multiscale superpixel graph construction, difference graph convolution, and multi-level fusion strategy. First, we preprocess the original image by using the full convolutional networks to extract superpixel-level features, where the visual primitive based on superpixel learned can be regarded as an independent clustering process. Second, we construct multiscale superpixel maps in a progressively merged manner. Then, perform central difference graph convolution extracted with multiscale superpixel as graph nodes. Finally, we carry on multi-level fusion rule formulation to align semantics for better adaptation to downstream tasks.

\subsection{Superpixel-level Representation}
A good visual representation should have three attributes, namely object-level, textual-aligned, and semantic-rich \cite{b28}. We hypothesize that an object-level representation should possess the following three key characteristics: finer-grained, robustness, and comprehensiveness. Intuitively, superpixel-level regions are more suitable for humans to perceive arbitrary shapes of complex objects in the real world. Compared with region or patch-level methods, superpixel is more robust in detecting edge features. Moreover, as an over-segmentation method, superpixel essentially preserves the down-sampling or up-sampling details, which helps to capture the spatial coherence in the local regions. Based on this strategy, we adopt superpixel as learnable compact visual inputs to provide a semantically meaningful tessellation of visual content. 

Traditional clustering-based superpixel algorithms (e.g., SLIC \cite{b29}) are difficult to implement on deep networks due to the non-differentiable of their nearest neighbor operation. Inspired by \cite{b30}, we adopt a differentiable clustering method that employs a soft association map to replace the hard association matrix in the existing methods. Thus, fully convolutional networks can be easily applied for implementation. To theoretically define the preliminary strategy, we first convert the input image $I\in\mathbb{R}^{H\times W}$ into $N$ regular grid cells, which are regarded as initial superpixel. Then, we employ a standard encoder-decoder framework with skip connections to calculate soft association map $Q\in\mathbb{R}^{H\times W\times N}$ between pixels and superpixels. Specifically, we set $x_{i,j}\in\mathbb{R}^{1\times D}$ be the flattened $D$-dimensional feature vector of pixel coordinate ($i$,$j$). Subsequently, the $x_{i,j}\in\mathbb{R}^{1\times D}$ are concatenated with the 2-$D$ positional pixel features $y_{i,j}\in\mathbb{R}^{1\times2}$ to control the spatial compactness of superpixel. Then, we feed the obtained pixel features $\delta\in\mathbb{R}^{H\times W\times(D+2)}$ into the encoder-decoder network to learn pixel-superpixel soft association $Q\in\mathbb{R}^{H\times W\times N}$ and superpixel centers. The center of any superpixel $S$ consists of the $u_s$ feature vector and $r_s$ location vector, which is given by
\begin{equation}
u_s=\frac{\sum_{(i,j)\in I} x_{i,j}\cdot q^{s}_{i, j}}{\sum_{(i,j)\in I} q^{s}_{i, j}}
\label{eq:1}      
\end{equation}
\begin{equation}
r_s=\frac{\sum_{(i,j)\in I} y_{i,j}\cdot q^{s}_{i, j}}{\sum_{(i,j)\in I} q^{s}_{i, j}}
\label{eq:2}      
\end{equation}
where the entry $q^{s}_{i, j}$ indicates the probability that pixel coordinate ($i$,$j$) is assigned to the $s$th superpixel, such that $q^{s}_{i, j}\in[0,1]$. Here, $\sum_{(i,j)\in I} q^{s}_{i, j}$ is the normalization constant.

Mathematically, $u_s$ and $r_s$ are the weighted sum of features and locations for all pixels. To this end, $Q$ can be normalized to make $\sum_{(i,j)\in I} q^{s}_{i, j}=1$. The superpixel center can be redefined as follows
\begin{equation}
\begin{bmatrix}
u_1 & r_1 \\
\vdots& \vdots \\
u_N & r_N
\end{bmatrix}
=\hat{Q}^{T}\delta
\label{eq:3}      
\end{equation}
where $\hat{Q}$ indicates the column-normalized $Q$, and $T$ is the matrix transpose.

In practice, it is time-consuming and expensive to compute all pixel-superpixel pairs. Therefore, we constrain its search region to the finite $|M_{i,j}|=9$ surrounding grid cells to speed up the sampling process, which reduces the size of $Q$ from $H\times W\times N$ to $H\times W\times9$. Thus, we have $Q\in\mathbb{R}^{H\times W\times|M_{i,j}|}$. Afterward, the reconstructed feature and location of each pixel can be derived by updating $Q$ and superpixel centers, that is

\begin{equation}
u_{i,j}=\frac{\sum^N_{s=1}(u_s\cdot q^s_{i,j})}{\sum^N_{s=1} q^s_{i,j}}=\frac{\sum_{s\in M_{i,j}}(u_s\cdot q^s_{i,j})}{\sum_{s\in M_{i,j}} q^s_{i,j}}
\label{eq:4}      
\end{equation}
\begin{equation}
r_{i,j}=\frac{\sum^N_{s=1}(r_s\cdot q^s_{i,j})}{\sum^N_{s=1} q^s_{i,j}}=\frac{\sum_{s\in M_{i,j}}(r_s\cdot q^s_{i,j})}{\sum_{s\in M_{i,j}} q^s_{i,j}}
\label{eq:5}      
\end{equation}

Further, Eq.~(\ref{eq:4}) and Eq.~(\ref{eq:5}) of superpixel centers can be rewritten by
\begin{equation}
\hat{\delta}=\tilde{Q}
\begin{bmatrix}
u_1 & r_1 \\
\vdots& \vdots \\
u_N & r_N
\end{bmatrix}
=\tilde{Q}\hat{Q}^{T}\delta
\label{eq:6}      
\end{equation}
where $\tilde{Q}$ denotes the normalized association matrix $Q$ satisfying $\sum_{s\in M_{i,j}} q^s_{i,j}=1$.

To encourage the superpixel generation model to group pixels with similar features of interest, the reconstruction loss function of the association matrix $Q$ is calculated as
\begin{equation}
Loss_{\text{rec}}=\text{dist}(\hat{\delta}, \delta)=\text{dist}(\tilde{Q}\hat{Q}^{T}\delta, \delta)
\label{eq:7}      
\end{equation}
where $\text{dist}(\cdot)$ is the distance metric.

Finally, we provide a compactness loss function, which enforces the superpixel to be spatially compact as follows
\begin{equation}
\begin{aligned}
\begin{split}
Loss_{\text{compact}}&=\frac{\sum^N_{s=1}\sum_{(i,j)\in I} {\parallel r_s-y_{i,j}}\parallel_2}{H\cdot W\cdot N} \\
&=\frac{\sum_{s\in M_{i,j}}\sum_{(i,j)\in I} {\parallel r_s-y_{i,j}}\parallel_2}{H\cdot W\cdot\mid M_{i,j}\mid}
\end{split}
\end{aligned}
\label{eq:8}      
\end{equation}
where $\parallel\cdot\parallel$ denotes the L2-norm.

\subsection{Multiscale Superpixel Graph Construction}
Generally, over-segmentation leads to a higher number of superpixels and a smaller generated volume. While each superpixel exhibits homogeneity, there is still a loss of spatial local information. As a result, the role of spatial information in classification is weakened. Considering contextual features at different scales helps to explore the spatial information of images more comprehensively, we design a multiscale graph structure by progressively merging adjacent superpixels. We first define a four-connected graph as $\mathcal{G}=(\mathcal{V},\varepsilon)$ consisting of nodes $v\in \mathcal{V}$ and edges $e\in\varepsilon\subseteq\mathcal{V}\times\mathcal{V}$ with cardinalities $n=\mid\mathcal{V}\mid$ and $m=\mid\varepsilon\mid$. Then, we assume an auxiliary forest with $n$ trees (i.e., $n$ nodes), where each superpixel is regarded as a node. Here, each edge $e_{i,j} =\parallel v_i$, $v_j \parallel_1$ is assigned a weight by using the $L_1$ distance to measure the dissimilarity between adjacent nodes. After that, the Boruvka algorithm \cite{b31} is applied to compute a minimum spanning tree (MST) in a bottom-up manner. Meanwhile, record the order in which each edge joins the MST. If an edge is added in the MST, the nodes $v_i$ and $v_j$ will be connected, which reduces the number of trees from $n$ to $n-1$. In this way, we can get superpixel maps of different scales from fine to coarse by repeating merging until $L$ trees are left in the auxiliary forest.

\begin{figure}[!t]
\centering
\includegraphics[width=2.65in]{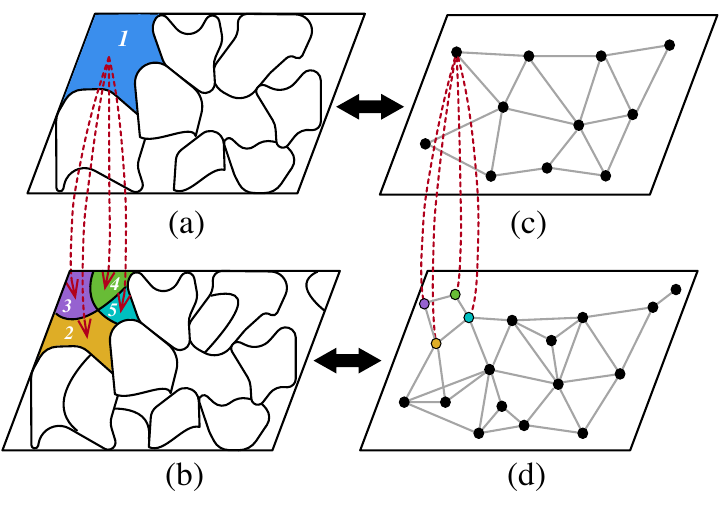}
\caption{Illustration of multi-scale superpixel graph construction.  (a) and (b) are large-scale and small-scale superpixel maps, respectively. Note that superpixel region $1$ in (a) is constructed by merging adjacent superpixels (region 2 to 5) in (b). (c) and (d) are corresponding generated graph nodes of (a) and (b), respectively.}
\label{fig_2}
\vspace*{-8pt}
\end{figure}
Different from the graph nodes representation at the pixel-level, incorporating superpixels into the graph significantly reduces the number of nodes and improves computational efficiency. In this work, we sequentially construct multiscale graphs with different numbers of nodes, as described in Fig.~\ref{fig_2}. For a structured graph with $L$ nodes, we have a multiscale segmentation graph as $\{S^{(k)}$, $\mathcal{G}^{(k)}\}^K_{k=1}$, where $K (K\ge1)$ is the number of superpixel maps. Then, we set $S^{(k)}=\{C^{(k)}_m\}^{L^{(k)}}_{m=1}$to be the $k$th superpixel maps consisting of $L^{(k)}$ superpixels. $\mathcal{G}^{(k)}=(\mathcal{V}^{(k)}$, $\mathcal{\varepsilon}^{(k)})$ formally denotes the graphs corresponding to $S^{(k)}$, $L^{(k)}(L^{(1)}>L^{(2)}>\dots>L^{(K)})$. Mathematically, the $\mathcal{V}^{(k)}$ and $\mathcal{\varepsilon}^{(k)}$ are generally defined by the node matrix $\textbf{H}^{(k)}$ and adjacency matrix $\textbf{A}^{(k)}$. Concretely, the $\textbf{\textit{i}}$th row of $\textbf{H}^{(k)}$ represents the $\textbf{\textit{i}}$th node in $V^{(k)}$, and the edge weight between the $\textbf{\textit{i}}$th and $\textbf{\textit{j}}$th nodes is represented as $\textbf{A}^{(k)}_\textbf{\textit{i, j}}$. Now, the adjacency matrix $\textbf{A}^{(k)}_\textbf{\textit{i, j}}$ of unweighted graph $\mathcal{G}^{(k)}$ can be formulated as
\begin{equation}
\textbf{A}^{(k)}_\textbf{\textit{i, j}}=
\begin{cases}
1, & \text{if}~C^{(k)}_\textbf{\textit{i}}~\text{and}~C^{(k)}_\textbf{\textit{j}} \text{are adjacent} \\
0, & \text{otherwise}
\end{cases}
\label{eq:9}      
\end{equation}
where $\textbf{A}^{(k)}_\textbf{\textit{i, j}}$ denotes the value of $\textbf{A}^{(k)}$ at location ($\textbf{\textit{i}}$, $\textbf{\textit{j}}$). Notably, two superpixel regions are considered adjacent if they share a common boundary.

However, assigning the same edge weights for all adjacent node pairs is unreasonable. Therefore, we will discuss the details of updating edge weights in \textit{Section~C}.

\subsection{Center Difference Graph Convolution}
Recently, graph convolutional networks (GCN) have achieved impressive performance due to conducting flexible convolution on non-structured graphs. Generally, vanilla graph convolution operations are directly aggregated node information in the graph topology. However, graph convolution lacks the ability to encode detailed gradient information, ignoring superpixel difference relations to achieve superpixel-level segmentation accurately. To mitigate the problems mentioned above, we present a center difference graph convolution (CDGCN) strategy to incorporate topological features. The CDGCN augments the vanilla graph convolution with a superpixel difference term while simultaneously learning local intensity and gradient information between center nodes and adjacent nodes, as well as features of nodes themselves to normalize edge weights. Remarkably, each pixel should be uniquely assigned to a superpixel, performing a difference graph convolution operation on the graph domain. 

The convolution operation on the vanilla graph is formulated as:
\begin{equation}
\textbf{H}^{(l+1)}=f_{\mathcal{G}}(\textbf{H}^{(l)}, \textbf{A})=\sigma(\textbf{A}\textbf{H}^{(l)}\textbf{W}^{(l)})
\label{eq:10}      
\end{equation}
where $\textbf{H}^{(l)}$ and $\textbf{H}^{(l+1)}$ denote the input of layer $l$th and the output matrix of layer ($l+1$)th, respectively, $\sigma$ denotes the non-linear activation function, and $\textbf{W}^{(l)}$ denotes the trainable weight matrix of layer $l$th. \textbf{A} is the normalized adjacency matrix with self-connections, expressed as
\begin{equation}
\textbf{A}=\hat{\textbf{D}}^{-\frac{1}{2}}\hat{\textbf{A}}\hat{\textbf{D}}^{-\frac{1}{2}}
\label{eq:11}      
\end{equation}
where $\hat{\textbf{A}}=\textbf{A}+\textbf{I}$ and $\hat{\textbf{D}}=\sum_{j}\hat{\textbf{A}}_{i,j}$.

Assuming a node $v_i$, the graph convolution centered at it is given by:
\begin{equation}
{\textbf{h}(v_i)}^{(l+1)}=\sigma\left(\sum\nolimits_{v_{j}\in{R}_{i}}\frac{1}{Z_{i,j}}{\textbf{h}(v_j)}^{(l)}\textbf{w}(\eta_{i}(v_j))^{(l)}\right)
\label{eq:12}      
\end{equation}
where $Z_{i,j}$ represents the normalized factor, $v_{j}^{(l)}$ represents all the adjacent node vectors of $v_i$ in layer $l$, i.e., $v_{j}\in{R}_{i}$. $\eta_i$ is a function \cite{b32} to divide nodes in ${R}_{i}$ into three subsets $d0$, $d1$, and $d2$. Formally, we have
\begin{equation}
\eta_{i}(v_j)=
\begin{cases}
d0, & \text{if}~r_j=r_i\\
d1, & \text{if}~r_j<r_i\\
d2, & \text{if}~r_j>r_i
\end{cases}
\label{eq:13}      
\end{equation}
where $d0$ denotes the node $v_i$ itself, $d1$ denotes the centripetal subset that the adjacent nodes are closer to the gravity center, $d2$ is the centrifugal subset that contains adjacent nodes away from the gravity center in $R_i$, and $r_i$ is the average distance.

The designed center difference operation can capture the local subtle differences of superpixels to enhance the representation and generalization capability. At the same time, it can effectively overcome the noise problem in superpixel segmentation. Thus, we use the center difference GCN to aggregate the center-oriented gradient, which is formulated as 
\begin{equation}
{\textbf{h}(v_i)}^{(l+1)}=\sum\nolimits_{v_{j}\in{R}_{i}}\frac{1}{Z_{i,j}}(\textbf{h}(v_j)-\textbf{h}(v_i))^{(l)}\textbf{w}(\eta_{i}(v_j))^{(l)}
\label{eq:14}      
\end{equation}
\begin{figure}[!t]
\centering
\includegraphics[width=3.55in]{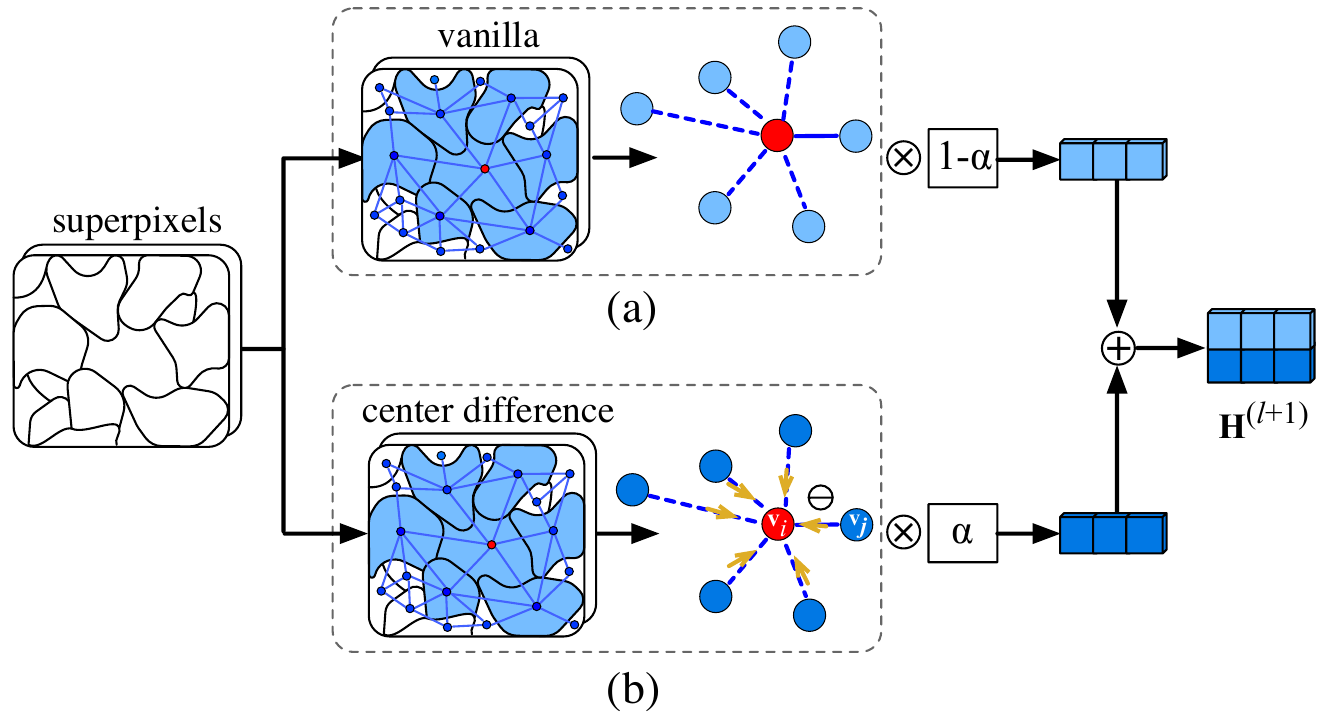}
\caption{Comparison of superpixel-level vanilla and difference graph convolutional layer. (a) Vanilla graph convolutional layer. (b) Difference graph convolutional layer. Where the minus signs denote the difference operation between the central node 
$v_i$ (red node) and adjacent node $v_j$ (blue node) in the blue sampling area at the superpixel-level. The yellow arrows denote the corresponding gradient features and their directions. The blue dashed and solid lines represent weak (or zero) and strong edge weights, respectively.}
\label{fig_3}
\vspace*{-8pt}
\end{figure}

Here, $\sigma$ is omitted to simplify the subsequent expression. It should be noted that Eq.~(\ref{eq:14}) mainly includes two steps of sampling and aggregation, but the aggregation step is different here. Next, we integrate Eq.~(\ref{eq:12}) and (\ref{eq:14}) to enhance robust and differentiated modeling capability. Here, a weight can be assigned to rewrite as 
\begin{equation}
\begin{aligned}
\begin{split}
{\textbf{h}(v_i)}^{(l+1)}&=\alpha\underbrace{\sum\nolimits_{v_{j}\in{R}_{i}}\frac{1}{Z_{i,j}}(\textbf{h}(v_j)-\textbf{h}(v_i))^{(l)}\textbf{w}(\eta_{i}(v_j))^{(l)}}_{\text{center difference term}}\\
&+(1-\alpha)\underbrace{\sum\nolimits_{v_{j}\in{R}_{i}}\frac{1}{Z_{i,j}}\textbf{h}(v_j)^{(l)}\textbf{w}(\eta_{i}(v_j))^{(l)}}_{\text{vanilla term}}
\end{split}
\end{aligned}
\label{eq:15}      
\end{equation}
where $\alpha\in[0,1]$ controls the assignment contribution between fine-grained gradient and nodes. The higher the $\alpha$ value, the more gradient information is included.

Further, Eq.~(\ref{eq:15}) can be redefined as follows
\begin{equation}
\begin{aligned}
\begin{split}
\textbf{H}^{(l+1)}&=\alpha\underbrace{(\textbf{A}\textbf{H}^{(l)}-\bar{\textbf{A}}\odot\textbf{H}^{(l)})\textbf{W}^{(l)}}_{\text{center difference term}}+(1-\alpha)\underbrace{\textbf{A}\textbf{H}^{(l)}\textbf{W}^{(l)}}_{\text{vanilla term}}\\
&=(\textbf{A}\textbf{H}^{(l)}-\alpha\bar{\textbf{A}}\odot\textbf{H}^{(l)})\textbf{W}^{(l)}
\end{split}
\end{aligned}
\label{eq:16}      
\end{equation}
where $\bar{\textbf{A}}$ represents the second dimension of the adjacency matrix $\textbf{A}$, i.e., $\sum_{j}\textbf{A}_{i,j}$, $\textbf{W}^{(l)}$ is shared between the two terms, and the operator $\odot$ is the Hadamard product.

Then, the superpixel-level center difference graph convolution is generalized to
\begin{equation}
\begin{bmatrix}
\mathcal{G}_{\gamma}(u_1)  \\
\vdots \\
\mathcal{G}_{\gamma}(u_N)
\end{bmatrix}
=\left(\textbf{A}
\begin{bmatrix}
u_1 \\
\vdots \\
u_N 
\end{bmatrix}
-\alpha\bar{\textbf{A}}\odot
\begin{bmatrix}
u_1 \\
\vdots \\
u_N 
\end{bmatrix}
\right)\textbf{W}
\label{eq:17}      
\end{equation}
where [$u_1$, \dots, $u_N$] is the feature set of the superpixels, and $\mathcal{G}_{\gamma}(u_{N})$ is the $N$th superpixel after $\gamma$ layers convolution operation. In this work, $\gamma=2$ is adopted.

By using Eq.~(\ref{eq:4}) to Eq.~(\ref{eq:6}), Eq.~(\ref{eq:17}) can be further updated as
\begin{equation}
\begin{aligned}
\begin{bmatrix}
\mathcal{G}_{\gamma}(u_1)  \\
\vdots \\
\mathcal{G}_{\gamma}(u_N)
\end{bmatrix}
&=\left(\textbf{A}\tilde{Q}\hat{Q}^{T}
\begin{bmatrix}
x_{1,1} & \cdots &x_{1,W}\\
\vdots   & \ddots &\vdots   \\
x_{H,j}  & \cdots &x_{H,W}
\end{bmatrix} \right.\\
& \left. -\alpha\bar{\textbf{A}}\odot\tilde{Q}\hat{Q}^{T}
\begin{bmatrix}
x_{1,1} & \cdots &x_{1,W}\\
\vdots   & \ddots &\vdots   \\
x_{H,j}  & \cdots &x_{H,W}
\end{bmatrix}
\right)\textbf{W} \\
&=(\textbf{A}\tilde{Q}\hat{Q}^{T}\delta_u-\alpha\bar{\textbf{A}}\odot\tilde{Q}\hat{Q}^{T}\delta_u)\textbf{W}
\end{aligned}
\label{eq:18}      
\end{equation}
where $\delta_{u}\in\mathbb{R}^{(H\times W\times D)}$ denotes the feature matrix, given as
\begin{equation}
\delta_u=
\begin{bmatrix}
x_{1,1} & \cdots &x_{1,W}\\
\vdots   & \ddots &\vdots   \\
x_{H,j}  & \cdots &x_{H,W}
\end{bmatrix}
\label{eq:19}      
\end{equation}

It should be noted that it is difficult for GCN to obtain the adjacency matrix and degree matrix by calculating the probability values of $q^{s}_{i, j}$. In fact, although the shape of the superpixel has changed, its relative positions are not affected, which will avoid recomputing the adjacency matrix $\textbf{A}$ and degree matrix $\textbf{D}$ when $Q$ is updated. Furthermore, features of the node itself and its directly adjacent are mapped in each execution, while other non-adjacent nodes are masked to better model the location information of superpixels. The process of our designed difference graph convolution strategy is shown in Fig.~\ref{fig_3}. To avoid overfitting with limited training samples, an average pool operation is performed for the images during the preprocessing. 
\begin{figure}[!t]
\centering
\includegraphics[width=3.4in]{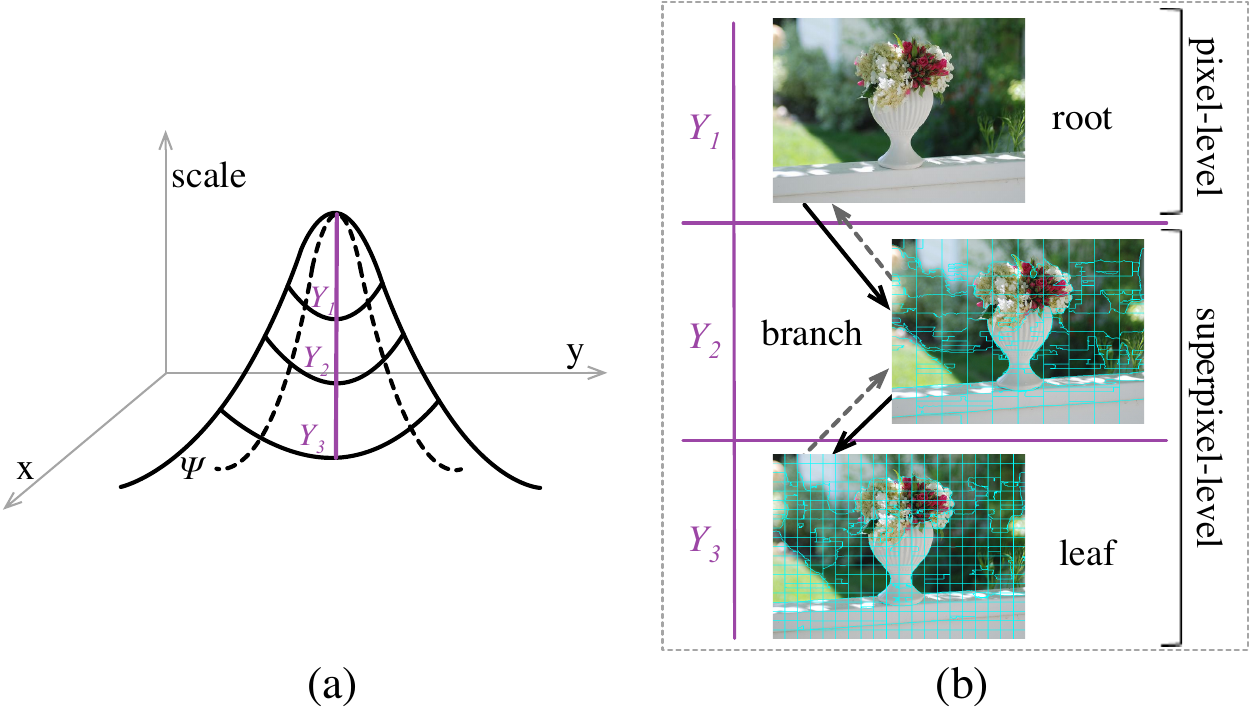}
\caption{Description of visual information at different spatial scales. (a) Built the scale space by convex surfaces $\Psi$, which can be quantized as a multi-level tree. (b) The process of the three-level parsing architecture, where the red dashed lines execute Tree-LSTM.}
\label{fig_4}
\end{figure}
\section{Multi-level Features Fusion}
By combining multiscale superpixels and difference GCNs for training under a unified framework, we can simultaneously capture the optimal parameters. While the spatial structure of images can be well modeled by superpixel-based difference GCN, it cannot fit the individual features of each pixel. In this work, we design a multi-level parsing architecture to integrate complementary features of different scales at pixel-level and superpixel-level into the image encoder, pursuing the hierarchical organization of visual content to facilitate understanding and reasoning about image structure. Different scales can present different levels of detail (i.e., “coarse to fine”), which helps to exploit the abundant local property of image regions from diverse levels. Therefore, we construct a large-scale graph and a small-scale graph with different numbers of nodes to enhance the discrimination of superpixel representations. In specific, image convolutions are first used to learn pixel-level local features. Note that we convert pixel features to image space by passing graph node features, which can avoid the structural incompatibility of the data between a CNN and GCN. Then, we build a three-level hierarchy framework from the pixel-level root to the branch layer of the superpixel-level large-scale and small-scale leaf layer, performing deeply structured modeling of images. Assuming that a convex surface $\Psi$, as described in Fig.~\ref{fig_4}(a). By quantizing along scales in $Y_j$, $j$ = ($1$, $2$, $\cdots$, $L$), we derive a tree structure by image hierarchical shape representation, which is demonstrated in Fig.~\ref{fig_4}(b). As a result, the constructed hierarchical tree can be denoted as $\mathcal{T}$= ($Y_1, Y_2, Y_3, \varepsilon_{tree}$), where $Y_1$ is defined as the root set to represent pixel-level features,  $Y_2=\{a_i\}_{i=1}^L$ and $Y_3=\{b_i\}_{i=1}^L$ define the branch set and leaf set of superpixel-level features, respectively, and $\varepsilon_{tree}$ is the connections. Taking inspiration from the Tree-LSTM \cite{b33} method for modeling image hierarchical structure in image captioning tasks, we utilize the Tree-LSTM to mine contextual information in a bottom-up manner to achieve feature enhancement. Here, the input vector of the pixel-level root node is set as the linear fusion of pixel-level mean-pooled features of branch ($\bar{a}=\frac{1}{L}\sum_{i=1}^{L}a_i$) and leaf ($\bar{b}=\frac{1}{L}\sum_{i=1}^{L}b_i$):$Y_1=W_{a}\bar{a}+W_{b}\bar{b}$, where $W$ are input weights matrices. At last, we output multi-level representations to enrich image features. 

\section{Experiments}
\subsection{Configuration Details}
For visual features, all input images are resized to 640$\times$640. Specifically, in the pixel-level spatial, the image convolution selects the ResNet-101 as the backbone. For the superpixel-level, we segment the image into two superpixel maps of different scales, 128$\times$128 and 64$\times$64, which are generated by a progressive merging manner. It should be noted that the pixels in each superpixel belong to the same category as much as possible. For graph convolutions, we adopt the same convolution kernel configuration as the pixel-level to ensure the consistency of the feature extractor architecture. Furthermore, we apply ITM, MLM, and MVM during pre-training tasks. For the visual backbone, SGD is chosen as an optimizer with a learning rate of $1\times10^{-2}$ and weight decay of $5\times10^{-4}$. For the language backbone, the  AdamW optimizer is fixed with a learning rate of $1\times10^{-4}$ and weight decay of $1\times10^{-2}$. The warm-up ratio is set to 1/3, and the number of training iterations is 1000. We initially used the WordPiece tokenizer based on BERT for text features. We pre-train our model for 30 epochs. Our code operation is implemented with Python 3.6 and PyTorch 1.7, which runs on 8 NVIDIA A100 GPUs.
\begin{table}[!t]
\centering
\renewcommand{\arraystretch}{1.3}
\setlength{\abovecaptionskip}{0.3cm}
\caption{Statistics of the different downstream tasks. \\Note that “$\blacktriangle$” denotes the Karpathy split \cite{b34}.}
\label{table_1}
\begin{IEEEeqnarraybox}[\IEEEeqnarraystrutmode\IEEEeqnarraystrutsizeadd{2pt}{2pt}]{l/c/c/c}
\IEEEeqnarrayrulerow[1pt]\\
\mbox{Task} ~& \mbox{Dataset} ~~& \mbox{Train Split} ~~&\mbox{Test Split} \\
\IEEEeqnarrayrulerow[0.5pt]\\
\raisebox{-7pt}[0pt][0pt]{Pre-training}  ~~& \mbox{VG} ~& \mbox{train} ~~& \mbox{-}  \\
~& \mbox{MSCOCO} ~~&  \mbox{train+restval$^{\blacktriangle}$} ~~&\mbox{-} \\
\IEEEeqnarrayrulerow[0.5pt]\\
\raisebox{-12pt}[0pt][0pt]{VQA}  ~~& \mbox{VQA v2} ~& \mbox{train+val} ~~& \mbox{test-dev/test-std}  \\
~& \mbox{VQA-CP v2} ~~&  \mbox{train} ~~&\mbox{test-std} \\
~& \mbox{GQA} ~~&  \mbox{train} ~~&\mbox{test-std} \\
\IEEEeqnarrayrulerow[0.5pt]\\
\mbox{VR} ~& \mbox{NLVR$^2$} ~~& \mbox{train} ~~&\mbox{dev/test-P} \\
\IEEEeqnarrayrulerow[0.5pt]\\
\mbox{VE} ~& \mbox{SNLI-VE} ~~& \mbox{train} ~~&\mbox{val/test} \\
\IEEEeqnarrayrulerow[0.5pt]\\
\raisebox{-7pt}[0pt][0pt]{Image-text Retrieval}  ~~& \mbox{MSCOCO} ~& \mbox{train+restval$^{\blacktriangle}$} ~~& \mbox{test$^{\blacktriangle}$}  \\
~& \mbox{Flickr30K} ~~&  \mbox{train} ~~&\mbox{test$^{\blacktriangle}$} \\
\IEEEeqnarrayrulerow[1pt]\\
\end{IEEEeqnarraybox}
\vspace*{-13pt}
\end{table}   
\begin{table*}[!t]
\centering
\renewcommand{\arraystretch}{1.3}
\setlength{\abovecaptionskip}{0.3cm}
\caption{State-of-the-art comparison results on the VQA v2 dataset.}
\label{table_2}
\begin{IEEEeqnarraybox}[\IEEEeqnarraystrutmode\IEEEeqnarraystrutsizeadd{1pt}{1pt}]{l/c/c/c/c/c/c/c/c/c/c/c/c/c}
\IEEEeqnarrayrulerow[1pt]\\
\raisebox{-9pt}[0pt][0pt]{Method}~~&\raisebox{-9pt}[0pt][0pt]{Backbone}~~~&\raisebox{-9pt}[0pt][0pt]{Visual Embed}~~~&\IEEEeqnarraymulticol{4}{t}{\mbox{Test-dev (\%)}}&&\IEEEeqnarraymulticol{5}{t}{\mbox{Test-std (\%)}}\\
\cmidrule[0.5pt]{4-7}\cmidrule[0.5pt]{9-14}
&&&\mbox{Y/N}~~~~~~&\mbox{Num}~~~~~~&\mbox{Other}~~~~~~&\mbox{Overall} &~~~&\mbox{Y/N}~~~~~~&\mbox{Num}~~~~~~&\mbox{Other}~~~~~~&\mbox{Overall} \\  
\IEEEeqnarrayrulerow[0.5pt]\\
\mbox{UpDn \cite{b11}}~~&\mbox{R101}~~~&\raisebox{-28pt}[0pt][0pt]{\mbox{Object}}~~~~&81.82~~~~~~&44.21~~~~~~&56.05~~~~~~&65.32&~~~&82.20~~~~~~&43.90~~~~~~&56.26~~~~~~&65.67\\
\mbox{ViLBERT \cite{b41}}~~&\mbox{R101}~~~&\raisebox{-29pt}[0pt][0pt]{\mbox{detector}}~~~~&-~~~~~~&-~~~~~~&-~~~~~~&70.55&~~~&-~~~~~~&-~~~~~~&-~~~~~~&70.92\\
\mbox{LXMERT \cite{b42}}~~&\mbox{R101}~~~&~~~&87.00~~~~~~&52.66~~~~~~&61.57~~~~~~&72.42&~~~&88.20~~~~~~&54.20~~~~~~&63.10~~~~~~&72.50\\                                                                         
\mbox{UNITER$_\textit{BASE}$\cite{b24}}~~&\mbox{R101}~~~&~~~~&-~~~~~~&-~~~~~~&-~~~~~~&72.70&~~~&-~~~~~~&-~~~~~~&-~~~~~~&72.91\\ 
\mbox{OSCAR$_\textit{BASE}$\cite{b25}}~~&\mbox{R101}~~~&~~~~&-~~~~~~&-~~~~~~&-~~~~~~&73.16&~~~&-~~~~~~&-~~~~~~&-~~~~~~&73.44\\ 
\mbox{VILLA$_\textit{BASE}$\cite{b26}}~~&\mbox{R101}~~~&~~~~&-~~~~~~&-~~~~~~&-~~~~~~&73.59&~~~&-~~~~~~&-~~~~~~&-~~~~~~&73.67\\ 
\mbox{MLVQA \cite{b43}}~~&\mbox{R101}~~~&~~~~&86.64~~~~~~&51.90~~~~~~&60.53~~~~~~&70.30&~~~&-~~~~~~&-~~~~~~&-~~~~~~&70.57\\ 
\IEEEeqnarrayrulerow[0.5pt]\\
\mbox{Pixel-BERT \cite{b6}}~~&\mbox{R50}~~~&\raisebox{-12pt}[0pt][0pt]{\mbox{Pixel}}~~~~&-~~~~~~&-~~~~~~&-~~~~~~&71.35&~~~&-~~~~~~&-~~~~~~&-~~~~~~&71.42\\ 
\mbox{SOHO \cite{b7}}~~&\mbox{R101}~~~&~~~~&-~~~~~~&-~~~~~~&-~~~~~~&73.25&~~~&-~~~~~~&-~~~~~~&-~~~~~~&73.47\\ 
\mbox{TRAR \cite{b44}}~~&\mbox{X152}~~~&~~~~&88.11~~~~~~&55.33~~~~~~&63.31~~~~~~&72.93&~~~&-~~~~~~&-~~~~~~&-~~~~~~&-\\ 
\IEEEeqnarrayrulerow[0.5pt]\\
\mbox{ViLT \cite{b8}}~~&\mbox{ViT-B/32}~~~&\raisebox{-6pt}[0pt][0pt]{\mbox{Patch}}~~~~&-~~~~~~&-~~~~~~&-~~~~~~&71.26&~~~&-~~~~~~&-~~~~~~&-~~~~~~&-\\
\mbox{PTP-ViLT \cite{b45}}~~&\mbox{ViT-B/16}~~~&~~~~&-~~~~~~&-~~~~~~&-~~~~~~&72.13&~~~&-~~~~~~&-~~~~~~&-~~~~~~&73.36\\ 
\IEEEeqnarrayrulerow[0.5pt]\\
\IEEEeqnarrayseprow[2pt]\\
\mbox{\textbf{Ours}}~~&\mbox{R101}~~~&\mbox{Superpixel}~~~~&\textbf{90.09}~~~~~~&\textbf{57.24}~~~~~~&\textbf{64.51}~~~~~~&\textbf{74.42}&~~~&\textbf{90.04}~~~~~~&\textbf{58.01}~~~~~~&\textbf{67.37}~~~~~~&\textbf{74.48}\\
\IEEEeqnarrayrulerow[1pt]\\
\end{IEEEeqnarraybox}
\vspace*{-10pt}
\end{table*} 
\begin{table}[!t]
\centering
\renewcommand{\arraystretch}{1.3}
\setlength{\abovecaptionskip}{0.3cm}
\caption{Performance evaluation with the state-of-the-art \\approaches on the VQA-CP v2.}
\label{table_3}   
\begin{IEEEeqnarraybox}[\IEEEeqnarraystrutmode\IEEEeqnarraystrutsizeadd{2pt}{2pt}]{l/c/c/c/c/c}
\IEEEeqnarrayrulerow[1pt]\\ 
\mbox{Method} ~~~~&\mbox{Y/N}~~~~&\mbox{Num} ~~~~&\mbox{Other}~~~~&\mbox{Overall}\\
\IEEEeqnarrayrulerow[0.5pt]\\
\mbox{MLVQA}\mbox{\cite{b43}}~~~~&40.84~~~~&13.24~~~~&48.83~~~~&41.08\\
\mbox{NSM}\mbox{\cite{b46}}~~~~&- ~~~~&- ~~~~ &- ~~~~ &45.80\\
\mbox{DecompLR}\mbox{\cite{b47}}~~~~&70.99~~~~&18.72~~~~&45.57~~~~&48.87\\
\mbox{VGQE}\mbox{\cite{b48}}~~~~&66.35~~~~&27.08~~~~&46.77~~~~&50.11\\ 
\mbox{LMH}\mbox{\cite{b49}}~~~~&72.95~~~~&31.90~~~~&47.79~~~~&52.73\\ 
\mbox{UpDn+LPF}\mbox{\cite{b50}}~~~~&88.61 ~~~~&23.78~~~~&46.57~~~~&55.34\\
\mbox{GGE-DQ$_{\textit{tog}}$}\mbox{\cite{b51}}~~~~&87.04~~~~&27.75~~~~&49.59~~~~&57.32\\
\IEEEeqnarrayrulerow[0.5pt]\\
\textbf{\mbox{Ours}}~~~~&\textbf{89.96} ~~~~&\textbf{38.35} ~~~~&\textbf{50.22} ~~~~&\textbf{58.04}\\
\IEEEeqnarrayrulerow[1pt]\\
\end{IEEEeqnarraybox}
\vspace*{-12pt}
\end{table} 

\begin{table}[!t]
\centering
\renewcommand{\arraystretch}{1.3}
\setlength{\abovecaptionskip}{0.3cm}
\caption{Comparison with the state-of-the-art approaches \\on the GQA.}
\label{table_4}   
\begin{IEEEeqnarraybox}[\IEEEeqnarraystrutmode\IEEEeqnarraystrutsizeadd{2pt}{2pt}]{l/c/c/c}
\IEEEeqnarrayrulerow[1pt]\\ 
\mbox{Method} ~~~&\mbox{Backbone}~~~~&\mbox{Test-dev (\%)}~~~~&\mbox{Test-std (\%)} \\
\IEEEeqnarrayrulerow[0.5pt]\\
\mbox{LXMERT}\mbox{\cite{b42}}~~~&\mbox{R101}~~~~&60.00 ~~~~&60.33 \\
\mbox{12-in-1}\mbox{\cite{b52}}~~~&\mbox{R101}~~~~&- ~~~~&60.65 \\
\mbox{VL-T5}\mbox{\cite{b53}}~~~&\mbox{R101}~~~~&- ~~~~&60.80\\
\mbox{OSCAR$_{\textit{BASE}}$}\mbox{\cite{b25}}~~~&\mbox{R101}~~~~&61.58~~~~&61.62\\
\mbox{MDETR}\mbox{\cite{b54}}~~~&\mbox{ENB5}~~~~&62.95~~~~&62.45\\
\mbox{NSM}\mbox{\cite{b46}}~~~&\mbox{R101}~~~~&62.95 ~~~~&63.17\\
\mbox{VinVL$_{\textit{BASE}}$}\mbox{\cite{b37}}~~~&\mbox{X152}~~~~&65.05~~~~&64.65\\
\IEEEeqnarrayrulerow[0.5pt]\\
\textbf{\mbox{Ours}}~~~&\mbox{R101}~~~~&\textbf{65.49} ~~~~&\textbf{64.87} \\
\IEEEeqnarrayrulerow[1pt]\\
\end{IEEEeqnarraybox}
\vspace*{-18pt}
\end{table} 
\begin{table*}[!t]
\centering
\renewcommand{\arraystretch}{1.3}
\setlength{\abovecaptionskip}{0.3cm}
\caption{Image-text retrieval results with the state-of-the-art methods on the MSCOCO (5K test set) dataset. \\\textit{LARGE}: model size. The best results are in bold.}
\label{table_5}
\begin{IEEEeqnarraybox}[\IEEEeqnarraystrutmode\IEEEeqnarraystrutsizeadd{1pt}{1pt}]{l/c/c/c/c/c/c/c/c/c}
\IEEEeqnarrayrulerow[1pt]\\
\raisebox{-9pt}[0pt][0pt]{Method}~~~&\raisebox{-9pt}[0pt][0pt]{Backbone}~~~~&\raisebox{-9pt}[0pt][0pt]{Parameters}~~~~&\IEEEeqnarraymulticol{3}{t}{\mbox{TR}}~~~~~~&\IEEEeqnarraymulticol{3}{t}{\mbox{IR}}\\
\cmidrule[0.5pt]{4-6}\cmidrule[0.5pt]{8-10}
&&&\mbox{R@1}~~~~~&\mbox{R@5}~~~~~&\mbox{R@10}~~&~~&\mbox{R@1}~~~~~&\mbox{R@5}~~~~~&\mbox{R@10} \\  
\IEEEeqnarrayrulerow[0.5pt]\\
\mbox{UNITER$_{\textit{LARGE}}$\cite{b24}}~~~&\mbox{R101}~~~~&\mbox{300M}~~~~&65.70~~~~~&88.60~~~~~&93.80~~&~~&52.90~~~~~&79.90~~~~~&88.00 \\ 
\mbox{PTP-ViLT\cite{b45}}~~~&\mbox{ViT-B/16}~~~~&\mbox{87M}~~~~&67.10~~~~~&90.50~~~~~&94.30~~&~~&45.30~~~~~&79.10~~~~~&88.40 \\
\mbox{VinVL$_{\textit{LARGE}}$\cite{b37}}~~~&\mbox{X152}~~~~&\mbox{550M}~~~~&75.40~~~~~&92.90~~~~~&96.20~~&~~&58.80~~~~~&83.50~~~~~&90.30\\ 
\mbox{METER-CLIP \cite{b15}}~~~&\mbox{ViT-B/16}~~~~&\mbox{380M}~~~~&76.20~~~~~&93.20~~~~~&96.80~~&~~&57.10~~~~~&82.70~~~~~&90.10 \\
\mbox{ALIGN \cite{b55}}~~~&\mbox{EfficientNet-L2}~~~~&\mbox{490M}~~~~&77.00~~~~~&93.50~~~~~&96.90~~&~~&59.90~~~~~&83.30~~~~~&89.80\\ 
\mbox{ALBEF (14M)\cite{b17}}~~~&\mbox{ViT-B/16}~~~~&\mbox{210M}~~~~&77.60~~~~~&94.30~~~~~&97.20~~&~~&60.70~~~~~&84.30~~~~~&90.50\\ 
\mbox{X-VLM (16M)\cite{b16}}~~~&\mbox{ViT-B/16}~~~~&\mbox{216M}~~~~&81.20~~~~~&95.60~~~~~&98.20~~&~~&63.40~~~~~&85.80~~~~~&91.50\\ 
\mbox{BLIP\cite{b18}}~~~&\mbox{ViT-L/16}~~~~&\mbox{220M}~~~~&82.40~~~~~&95.40~~~~~&97.90~~&~~&65.10~~~~~&86.30~~~~~&91.80\\   
\IEEEeqnarrayrulerow[0.5pt]\\
\mbox{\textbf{Ours}}~~~&\mbox{R101}~~~~&\mbox{255M}~~~~&\textbf{82.60}~~~~~&\textbf{95.80}~~~~~&\textbf{97.95}~~&~~&\textbf{65.40}~~~~~&\textbf{87.10}~~~~~&\textbf{91.90} \\ 
\IEEEeqnarrayrulerow[1pt]\\
\end{IEEEeqnarraybox}
\vspace*{-4pt}
\end{table*} 

\subsection{Pre-training Datasets}
We built the same “in-domain” datasets as in UNITER\cite{b24} for our pretraining. It consists of the MSCOCO\cite{b35} and VG for four popular VL downstream tasks, encompassing VQA, VR, VE, and image-text retrieval. Different datasets are summarized in Table \ref{table_1}.
VQA requires inferring text questions related to the contents of the images, and we evaluate VQA on publicly available VQA v2\footnote{https://visualqa.org/download.html} and VQA-CP v2\footnote{https://computing.ece.vt.edu/$\sim$aish/vqacp/} datasets. Specifically, VQA v2 is an extended version of VQA v1, which contains about 82.8K images and 443.8K questions for the \textit{train} set, 40.5K images, and 214.4K questions for the \textit{val} set. All questions are classified into three types: “\textit{Yes/No}”, “\textit{Number}”, and “\textit{Other}”. VQA-CP v2 dataset is reorganized from VQA v2, which has changed the prior distributions of answers to overcome the question-oriented bias problem. It is divided into \textit{train} and \textit{test} sets, containing about 121K images and 438K questions, 98K images, and 220K questions, respectively. Every image-text pair in VQA v2 and VQA-CP v2 is answered by 10 independent human annotators. An accuracy-based evaluation metric is defined to predict \textit{ans} as follows:
\begin{equation}
accuracy(ans)=min\left\{\frac{N(ans)}{3}, 1\right\}
\label{eq:20}      
\end{equation}
where $N$($\textit{ans}$) is the number of ans that humans provided answers by different annotators. Compared with VQA datasets, GQA\footnote{https://cs.stanford.edu/people/dorarad/gqa/} involves multistep reasoning and fewer language biases. GQA contains about 113K images and 22M questions. Here, we use GQA to evaluate the visual understanding ability of our model.

Unlike the VQA task, VR focuses more on reasoning relations and quantities. In our work, the task of VR is conducted on the NLVR$^2$ dataset\footnote{https://lil.nlp.cornell.edu/nlvr/}. It contains 103.2K, 8.2K images, and 86.4K, 7.0K questions for \textit{train} and \textit{dev} sets, respectively. For VE, it is used as a fine-grained visual reasoning task to infer whether the image semantically entails text. We perform the VE task on the SNLI-VE\footnote{https://github.com/necla-ml/SNLI-VE}, constructed based on Flickr30K\cite{b36} and Stanford Natural Language Inference (SNLI) datasets. SNLI-VE is split into \textit{train}, \textit{val}, and \textit{test} sets. Each of them consists of 29.8K, 1K, and 1K images with 529.5K, 17.9K, and 17.9K questions, respectively. Image-text retrieval typically requires models to retrieve the most relevant captions from candidate images. It consists of two subtasks, image-to-text retrieval (TR) and text-to-image retrieval (IR). Here, the performance of our proposed model is evaluated on the Flickr30K and MSCOCO datasets. Flickr30K dataset is split into train and test. The two splits consist of about 29K and 1K images with approximately 145K and 5K questions, respectively. The Recall@K (K=1,5,10) is conducted as the measure metrics in our experiments. Where R@K denotes the percentage of correct matchings in the top-K lists. The higher R@K shows better performance.

\begin{table*}[!t]
\centering
\renewcommand{\arraystretch}{1.3}
\setlength{\abovecaptionskip}{0.3cm}
\caption{Image-text retrieval comparison results with the state-of-the-art methods \\on the Flickr30K (1K test set) dataset.}
\label{table_6}
\begin{IEEEeqnarraybox}[\IEEEeqnarraystrutmode\IEEEeqnarraystrutsizeadd{1pt}{1pt}]{l/c/c/c/c/c/c/c/c/c}
\IEEEeqnarrayrulerow[1pt]\\
\raisebox{-9pt}[0pt][0pt]{Method}~~~~~~~&\raisebox{-9pt}[0pt][0pt]{Backbone}~~~~~~&\raisebox{-9pt}[0pt][0pt]{Parameters}~~~~&\IEEEeqnarraymulticol{3}{t}{\mbox{TR}}&~~~&\IEEEeqnarraymulticol{3}{t}{\mbox{IR}}\\
\cmidrule[0.5pt]{4-6}\cmidrule[0.5pt]{8-10}
&&&\mbox{R@1}~~~~~&\mbox{R@5}~~~~~&\mbox{R@10}~~&~~~&\mbox{R@1}~~~~~&\mbox{R@5}~~~~~&\mbox{R@10} \\  
\IEEEeqnarrayrulerow[0.5pt]\\ 
\mbox{ViLT \cite{b8}}~~~~~~~&\mbox{ViT-B/32}~~~~~~&\mbox{87M}~~~~&83.50~~~~~&96.70~~~~~&98.60~~&~~~&64.40~~~~~&88.70~~~~~&93.80\\ 
\mbox{PTP-ViLT \cite{b45}}~~~~~~~&\mbox{ViT-B/16}~~~~~~&\mbox{87M}~~~~&85.20~~~~~&96.90~~~~~&98.50~~&~~~&68.80~~~~~&91.40~~~~~&95.30\\
\mbox{UNITER$_{\textit{LARGE}}$\cite{b24}}~~~~~~~&\mbox{R101}~~~~~~&\mbox{300M}~~~~&87.30~~~~~&98.00~~~~~&99.20~~&~~~&75.60~~~~~&94.10~~~~~&96.80\\ 
\mbox{ViLBERT \cite{b41}}~~~~~~~&\mbox{R101}~~~~~~&\mbox{274M}~~~~&-~~~~~&-~~~~~&-~~&~~~&58.20~~~~~&84.90~~~~~&91.50\\  
\mbox{Pixel-BERT \cite{b6}}~~~~~~~&\mbox{X152}~~~~~~&\mbox{144M}~~~~&86.50~~~~~&98.10~~~~~&99.30~~&~~~&72.50~~~~~&92.70~~~~~&96.10\\ 
\mbox{VISTA$_{\textit{LARGE}}$\cite{b38}}~~~~~~~&\mbox{ViT-B/16}~~~~~~&-~~~~&89.50~~~~~&98.40~~~~~&99.60~~&~~~&75.80~~~~~&94.20~~~~~&96.90\\ 
\IEEEeqnarrayrulerow[0.5pt]\\
\mbox{\textbf{Ours}}~~~~~~~&\mbox{R101}~~~~~~&\mbox{255M}~~~~&\textbf{90.20}~~~~~&\textbf{98.60}~~~~~&\textbf{99.80}~~&~~~&\textbf{79.40}~~~~~&\textbf{94.50}~~~~~&\textbf{97.10} \\ 
\IEEEeqnarrayrulerow[1pt]\\
\end{IEEEeqnarraybox}
\vspace*{-8pt}
\end{table*} 

\subsection{Downstream Tasks Comparison Results}
In this section, we evaluate the effectiveness of our model on four downstream VL tasks. 

\textit{1) Visual question answering:} We formulate VQA as a multi-answer classification task and report the evaluation results on VQA v2, VQA-CP v2, and GQA benchmark datasets. Specifically, Table \ref{table_2} shows the experiment results on the VQA v2 dataset and compares the \textit{test-dev} and \textit{test-std} results to other state-of-the-art works. From Table \ref{table_2}, we can find that our model integrates superpixel as visual embedding, resulting in better performance compared to other works. In Table \ref{table_3}, we also evaluate our method on the VQA-CP v2 dataset, which is used to overcome the question-oriented bias. Note that ResNet 101 is used as the based backbone. Obviously, we achieve better performance in each question category compared to recent existing methods. Table \ref{table_4} provides the comparison results on the GQA, we can see that our method improves accuracy on all metrics. For example, we improve the overall accuracy of the VinVL \cite{b37} from 65.05\% to 65.49\% on the \textit{test-dev} set. 
\begin{figure}[ht]
\centering
\includegraphics[width=2.7in]{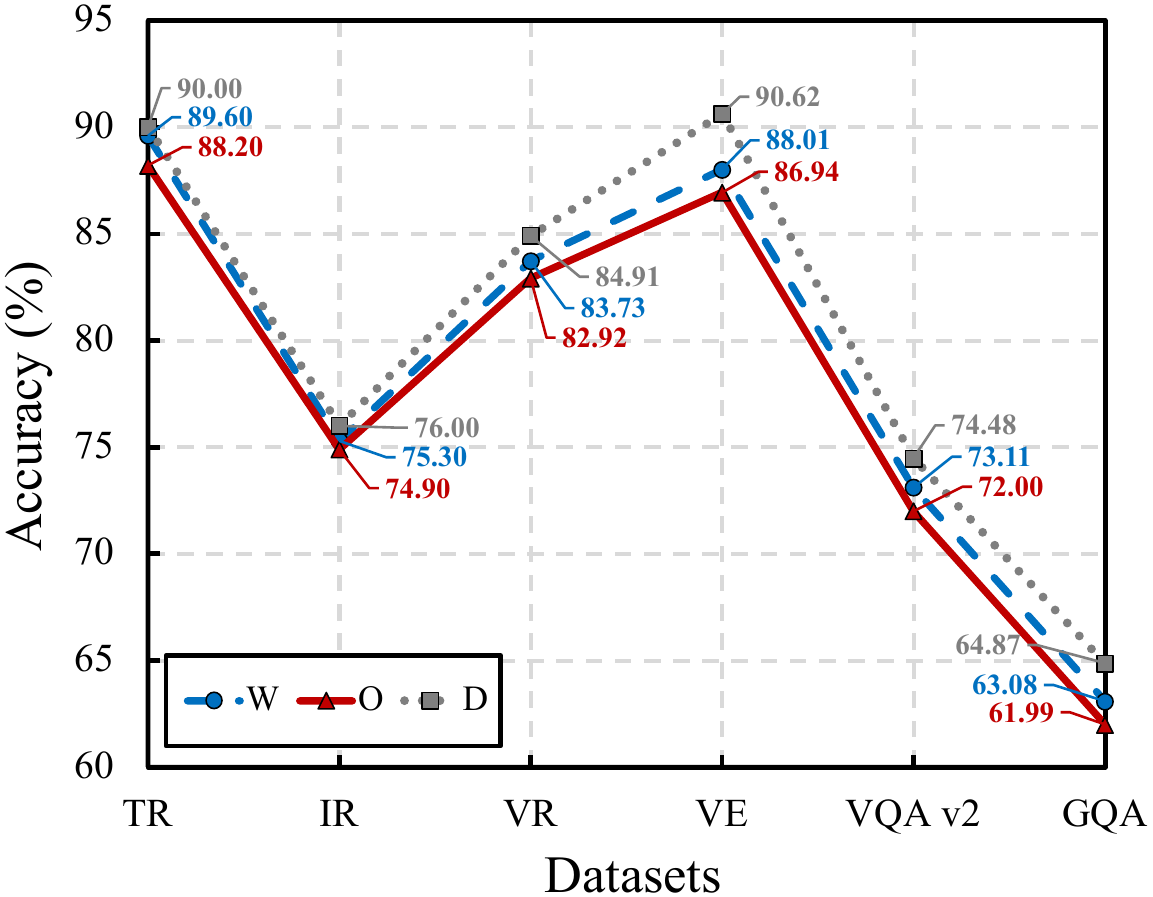}
\caption{Performance evaluation of the model in three states under different VL tasks. Here, “W” and “O” represent the model with and without GCN, respectively. “W+D” means that the model performs a difference convolution operation.}
\label{fig_5}
\vspace*{-8pt}
\end{figure}
\begin{figure}[ht]
\centering
\includegraphics[width=2.7in]{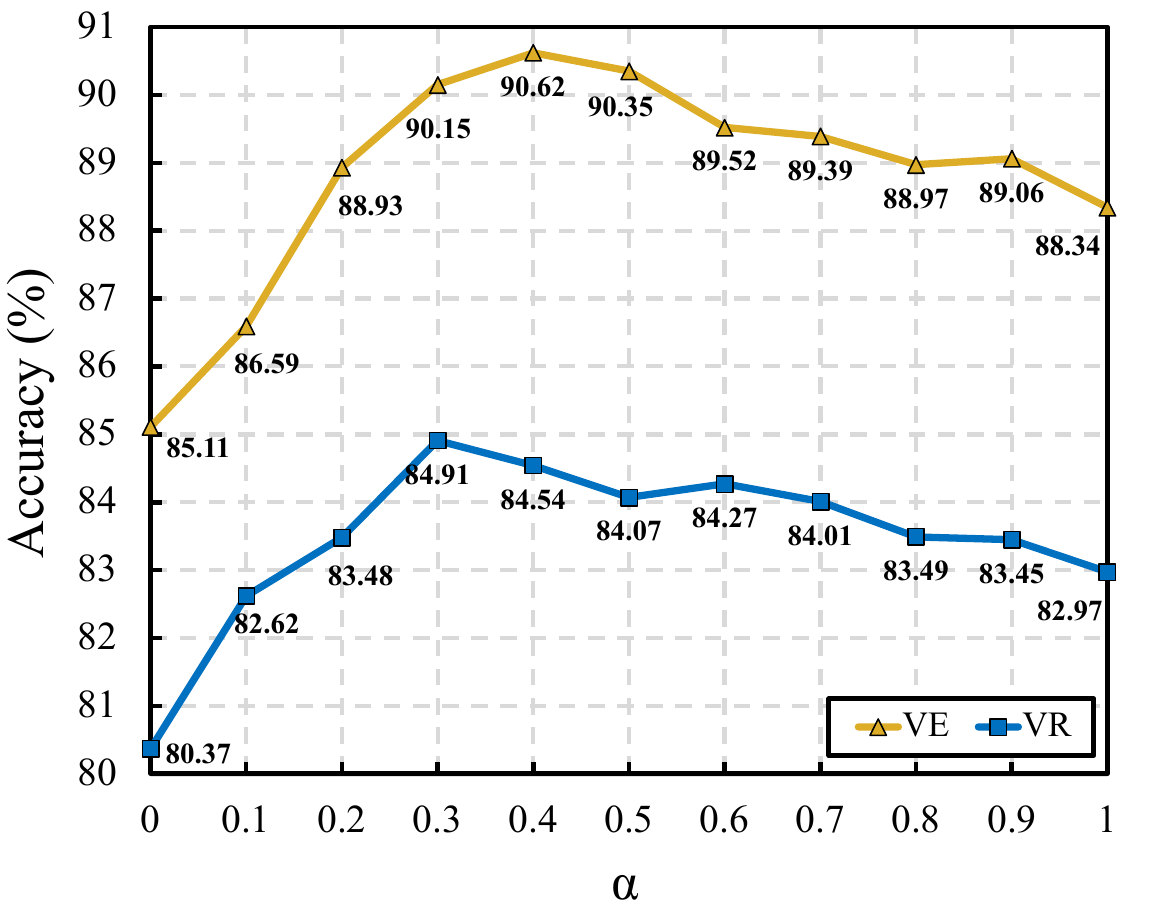}
\caption{Impact of different hyperparameter $\alpha$ on the accuracy of CDGC by following \textit{test} on SNLI-VE and \textit{dev} on NLVR$^2$ datasets. }
\label{fig_6}
\vspace*{5pt}
\end{figure}

\textit{2) Image-text retrieval:} As shown in Table \ref{table_5} and Table \ref{table_6}, we report competitive results on both TR and IR settings on MSCOCO and Flickr30K datasets, respectively. In addition, we also list the comparison of the basic backbone and pre-training parameters of different models.  In more detail, we outperform the best baseline BLIP \cite{b18} by 0.83\% on MSCOCO. Likewise, we outperform VISTA\cite{b38} by 0.81\% on Flickr30K, which fully proves the effectiveness of aligning VL features. 

\textit{3) Visual Reasoning:} VR is formulated as a binary classification task, which can address the problem of compositional reasoning on quantities, relations, and comparisons. In Table \ref{table_7}, we cconduct experiments on the NLVR$^2$ dataset, which consists of the \textit{dev} and \textit{test-P} sets. We achieve higher accuracy and bring 0.33\% gains on dev split than mPLUG \cite{b27}, which adopts a ViT-B/16-based visual backbone. 

\textit{4) Visual Entailment:} VE pursues visual reasoning tasks that are more fine-grained than VR and VQA. We evaluate the performance of our method and other excellent works on the SNLI-VE dataset. Data results are reported in Table \ref{table_8}. We find significant improvements over SimVLM \cite{b39} with huge size, which leads to a 4\% improvement on the \textit{dev} split. Moreover, our accuracy is even better than Prompt Tuning \cite{b40} on both \textit{dev} and \textit{test} sets. 

\begin{table}[!t]
\centering
\renewcommand{\arraystretch}{1.3}
\setlength{\abovecaptionskip}{0.3cm}
\caption{VR comparison results with the state-of-the-art \\methods on the NLVR$^2$ dataset.}
\label{table_7}   
\begin{IEEEeqnarraybox}[\IEEEeqnarraystrutmode\IEEEeqnarraystrutsizeadd{2pt}{2pt}]{l/c/c/c}
\IEEEeqnarrayrulerow[1pt]\\ 
\mbox{Method} ~~~~~~&\mbox{Backbone}~~~~~~~&\mbox{dev} ~~~~~~~&\mbox{test-P}\\
\IEEEeqnarrayrulerow[0.5pt]\\
\mbox{VILLA$_{\textit{LARGE}}$}\mbox{\cite{b26}}~~~~~~&\mbox{R101}~~~~~~~&79.76~~~~~~~&81.47\\ 
\mbox{OSCAR$_{\textit{LARGE}}$}\mbox{\cite{b25}}~~~~~~&\mbox{R101} ~~~~~~~&79.12~~~~~~~&80.37\\
\mbox{METER-CLIP}\mbox{\cite{b15}}~~~~~~&\mbox{ViT-B/16}~~~~~~~&82.33~~~~~~~&83.05\\
\mbox{ALBEF (14M)}\mbox{\cite{b17}}~~~~~~&\mbox{ViT-B/16}~~~~~~~&82.55~~~~~~~&83.14\\
\mbox{VinVL$_{\textit{LARGE}}$}\mbox{\cite{b37}}~~~~~~&\mbox{X152}~~~~~~~&82.67~~~~~~~&83.98\\
\mbox{X-VLM (16M)}\mbox{\cite{b16}}~~~~~~&\mbox{ViT-B/16}~~~~~~~&84.41~~~~~~~&84.76\\
\mbox{SimVLM$_{\textit{LARGE}}$}\mbox{\cite{b39}}~~~~~~&\mbox{ViT-B/16}~~~~~~~&84.13~~~~~~~&84.84\\
\mbox{mPLUG}\mbox{\cite{b27}}~~~~~~&\mbox{ViT-B/16} ~~~~~~~&84.58 ~~~~~~~ &84.95 \\
\IEEEeqnarrayrulerow[0.5pt]\\
\mbox{\textbf{Ours}}~~~~~~&\mbox{R101}~~~~~~~&\textbf{84.91}~~~~~~~&\textbf{84.98}\\
\IEEEeqnarrayrulerow[1pt]\\
\end{IEEEeqnarraybox}
\vspace*{-16pt}
\end{table}
\begin{table}[!h]
\centering
\renewcommand{\arraystretch}{1.3}
\setlength{\abovecaptionskip}{0.3cm}
\caption{VE comparison results with the state-of-the-art \\methods on the SNLI-VE dataset.}
\label{table_8}   
\begin{IEEEeqnarraybox}[\IEEEeqnarraystrutmode\IEEEeqnarraystrutsizeadd{2pt}{2pt}]{l/c/c/c}
\IEEEeqnarrayrulerow[1pt]\\ 
\mbox{Method} ~~~~~~~&\mbox{Backbone}~~~~~~~&\mbox{dev} ~~~~~~~&\mbox{test}\\
\IEEEeqnarrayrulerow[0.5pt]\\
\mbox{ALBEF (14M)}\mbox{\cite{b17}}~~~~~~~&\mbox{ViT-B/16}~~~~~~~&80.80~~~~~~~&80.91\\
\mbox{METER-CLIP}\mbox{\cite{b15}}~~~~~~~&\mbox{ViT-B/16}~~~~~~~&80.86~~~~~~~&81.19\\
\mbox{VILLA$_{\textit{LARGE}}$}\mbox{\cite{b26}}~~~~~~~&\mbox{R101}~~~~~~~&80.18~~~~~~~&80.02\\
\mbox{SOHO}\mbox{\cite{b7}}~~~~~~~&\mbox{R101}~~~~~~~&85.00~~~~~~~&84.95\\
\mbox{SimVLM$_{\textit{HUGE}}$}\mbox{\cite{b39}}~~~~~~~&\mbox{ViT-B/16}~~~~~~~&86.21~~~~~~~&86.32\\
\mbox{OFA$_{\textit{BASE}}$}\mbox{\cite{b56}}~~~~~~~&\mbox{R101} ~~~~~~~&89.30 ~~~~~~~ &89.20 \\
\mbox{mPLUG}\mbox{\cite{b27}}~~~~~~~&\mbox{ViT-B/16}~~~~~~~&89.45~~~~~~~&89.29\\
\mbox{Prompt Tuning}\mbox{\cite{b40}}~~~~~~~&R152~~~~~~~&90.04~~~~~~~&90.12\\
\IEEEeqnarrayrulerow[0.5pt]\\
\mbox{\textbf{Ours}}~~~~~~~&\mbox{R101}~~~~~~~&\textbf{90.21}~~~~~~~&\textbf{90.62}\\
\IEEEeqnarrayrulerow[1pt]\\
\end{IEEEeqnarraybox}
\vspace*{-21pt}
\end{table}  

\begin{figure*}[!t]
\centering
\includegraphics[width=6.2in]{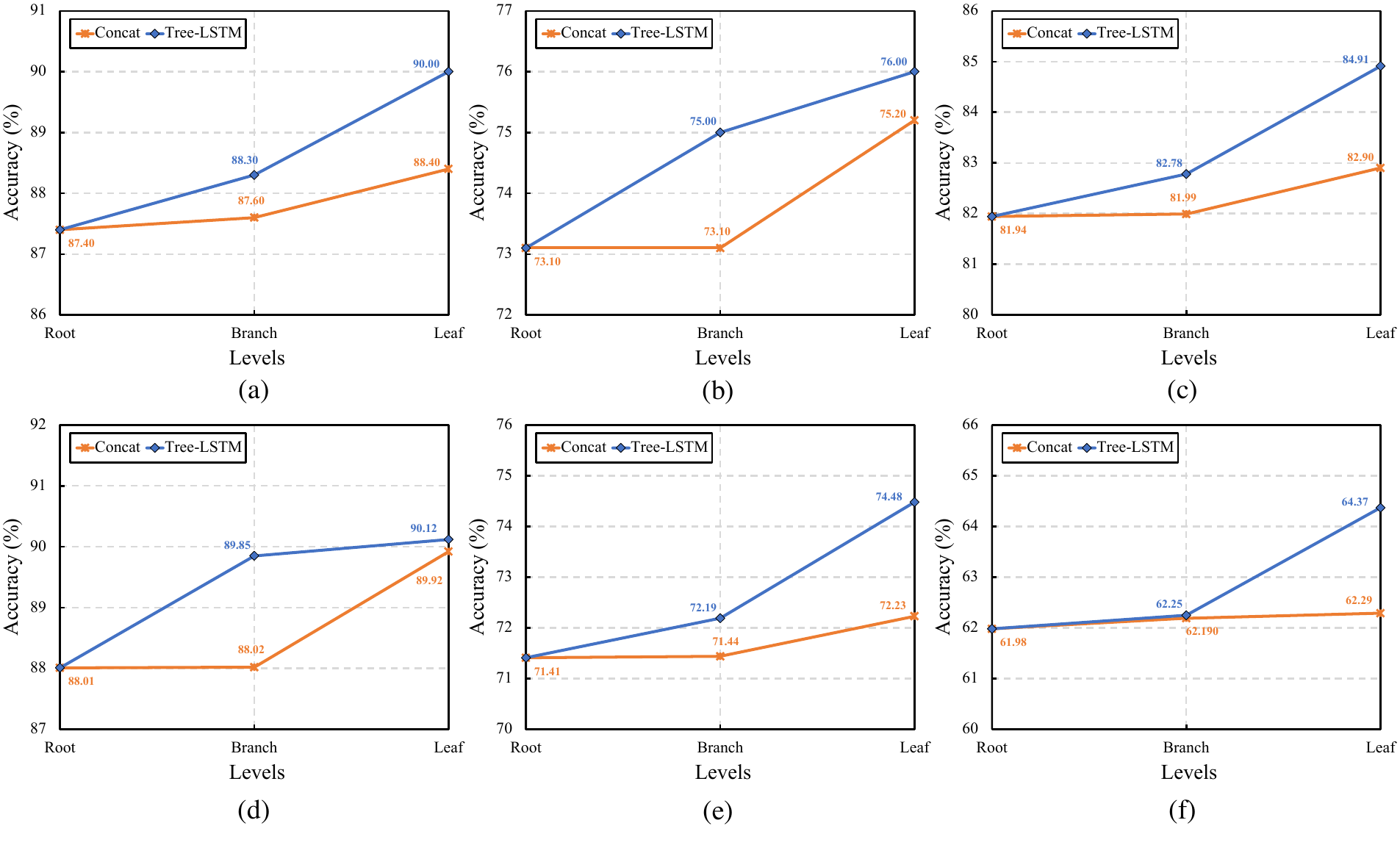}
\caption{Performance evaluation of the multi-level parsing architecture on multiple VL tasks. Here, (a) and (b) test TR@1 and IR@1 on MSCOCO (1K Test), respectively. (c) tests \textit{dev} on NLVR$^2$. (d) provides a \textit{test} on SNLI-VE. (e) and (f) give \textit{test-std} on VQA v2 and GQA. Note that “Concat” means the commonly used concatenation operation.}
\label{fig_7}
\end{figure*}
\subsection{Ablation Study}
In this work, we conduct several ablation studies to verify the contribution of each module.
 
\textit{1) Effectiveness of superpixel:} We focus on improving model performance by integrating superpixel features. In order to reflect the advantages of the superpixel features in our designed model, we conducted comparative experiments with different VL tasks at the pixel-level and the superpixel-level in Table \ref{table_9}. In addition, we also compare the pre-training time per epoch and memory to further examine the effectiveness of using superpixel. Here, the storage parameters of superpixel are obviously less than the pixel, which mainly includes the following two reasons. On the one hand, using pixels as graph nodes will lead to an increase in the number of parameters in the backbone stage, thereby increasing the computational cost. On the other hand, we store the adjacency matrix corresponding to the superpixel locally, which greatly alleviates the memory burden. 

\textit{2) Comparison of GCN and difference GCN:} We evaluate the impact of graph convolution and difference graph convolution on our model architecture. As shown in Fig.~\ref{fig_5}, we separately verify the performance comparison of the three states, “W”, “O”, and “W+D”, respectively. Specifically, “W” indicates that the model performs GCN. “O” indicates the proposed model without GCN. “W+D” means that the model implements a difference graph convolution operation. We conduct different VL downstream tasks. For image-text retrieval, our experiment is implemented by computing R@1 on the MSCOCO (1K test set) dataset. For NLVR$^2$, SNLI-VE, VQA v2, and GQA datasets, we use dev, test, test-std, and test-std respectively for evaluation. Based on the aforementioned observation, we can find that “W+D” achieves higher accuracy compared to the other two conditions, emphasizing the significant role of the difference process in intramodality modeling. 

\textit{3) Impact of $\alpha$ in CDGC:} The impact of different $\alpha$ values is verified in Fig.~\ref{fig_6}. We employ hyperparameter $\alpha$ to control the contribution of the spatial gradient cues in CDGC. Here, $\alpha$ is set to a learnable parameter value ranging from 0 to 1. As discussed in Eq.~(\ref{eq:16}), the higher the $\alpha$ value, the more critical the superpixel difference is, and vice versa. For the SNLI-VE dataset, the best performance is obtained at $\alpha=0.4$ (Accuracy = 90.62\%). Furthermore, CDGC also obtains the best accuracy when $\alpha=0.3$ (Accuracy = 84.91\%) on the VR dataset, which is better than the vanilla term (i.e., $\alpha=0$). This also fully proves that integrating superpixel difference operations into GCN is an effective solution.

\textit{4) Impact of different backbones:} Table \ref{table_10} tests the pre-training times and performance of our model on different backbones. As the number of network layers deepens, the increase in the number of parameters hinders computational efficiency. In particular, GCNs generally contain more parameters than traditional neural networks. To this end, we choose ResNet 101 as our backbone to improve computational efficiency. 

\textit{5) Impact of multi-level architecture:} We explore if multi-level parsing architecture helps in our work. In Fig.~\ref{fig_7}, we evaluate the performance of each level on multiple downstream VL tasks. As can be seen, using Tree LSTM at the branch level improves the accuracy remarkably, which is better than the general concatenation manner. Likewise, the leaf level with the Tree LSTM achieves higher accuracy. It is proven that the constructed three-level architecture can effectively capture intramodality features.
\begin{table*}[!t]
\centering
\renewcommand{\arraystretch}{1.3}
\setlength{\abovecaptionskip}{0.3cm}
\caption{Performance comparison of embed features with different VL tasks.}
\label{table_9}
\begin{IEEEeqnarraybox}[\IEEEeqnarraystrutmode\IEEEeqnarraystrutsizeadd{1pt}{1pt}]{l/c/c/c/c/c/c/c/c/c}
\IEEEeqnarrayrulerow[1pt]\\
\raisebox{-9pt}[0pt][0pt]{Embed Feature}~~~&\raisebox{-3pt}[0pt][0pt]{Training}~~&\raisebox{-3pt}[0pt][0pt]{Epoch}~~&\raisebox{-3pt}[0pt][0pt]{Memory}~~~&\IEEEeqnarraymulticol{2}{t}{\mbox{MSCOCO (1K test)}}~~&\mbox{NLVR$^2$}~~&\mbox{SNLI-VE}~~&\mbox{VQA v2}~~&\mbox{GQA}\\
\cmidrule[0.5pt]{5-10}
&\raisebox{2pt}[0pt][0pt]{Epochs}~~&\raisebox{2pt}[0pt][0pt]{Times (h)}~~&\raisebox{2pt}[0pt][0pt]{(MB)}~~&\mbox{TR@1}~~&\mbox{IR@1}~~&\mbox{dev}~~&\mbox{test}~~&\mbox{test-std}~~&\mbox{test-std} \\  
\IEEEeqnarrayrulerow[0.5pt]\\ 
\mbox{Pixel}~~~&30~~&4.3~~&3722~~~&88.20~~&74.00~~&82.19~~&86.06~~&71.92~~&62.01 \\ 
\mbox{Superpixel}~~~&30~~&3.1~~&1928~~~&90.00~~&76.00~~&84.91~~&90.12~~&74.48~~&64.37 \\ 
\IEEEeqnarrayrulerow[1pt]\\
\end{IEEEeqnarraybox}
\vspace*{-7pt}
\end{table*} 

\begin{table}[!t]
\centering
\renewcommand{\arraystretch}{1.3}
\setlength{\abovecaptionskip}{0.3cm}
\caption{Performance comparison of different backbones\\ in several VL tasks.}
\label{table_10}
\begin{IEEEeqnarraybox}[\IEEEeqnarraystrutmode\IEEEeqnarraystrutsizeadd{1pt}{1pt}]{l/c/c/c/c/c/c/c}
\IEEEeqnarrayrulerow[1pt]\\
\raisebox{-14pt}[0pt][0pt]{Backbone}&\raisebox{-7pt}[0pt][0pt]{Pre-training}&\IEEEeqnarraymulticol{2}{t}{\mbox{MSCOCO}}&\raisebox{-7pt}[0pt][0pt]{NLVR$^2$}&\raisebox{-7pt}[0pt][0pt]{SNLI-VE}&\raisebox{-7pt}[0pt][0pt]{VQA v2}&\raisebox{-7pt}[0pt][0pt]{GQA}\\
&\raisebox{-8pt}[0pt][0pt]{Times (h)}&\IEEEeqnarraymulticol{2}{t}{\mbox{(1K test)}}&&&& \\ 
\cmidrule[0.5pt]{3-9}
&&\mbox{TR@1}&\mbox{IR@1}&\mbox{dev}&\mbox{test}&\mbox{test-std}&\mbox{test-std} \\  
\IEEEeqnarrayrulerow[0.5pt]\\ 
\mbox{R18}&63&87.90&74.26&81.77&86.23&72.37&62.69 \\ 
\mbox{R101}&155&90.00&76.00&84.91&90.12&74.48&64.37 \\ 
\mbox{R152}&198&90.10&76.50&84.90&90.13&74.59&65.12 \\ 
\IEEEeqnarrayrulerow[1pt]\\
\end{IEEEeqnarraybox}
\vspace*{-14pt}
\end{table} 

\section{Conclusion}
This paper is the first to view superpixel as compact visual primitives. Superpixel is closer to the “reality” of human perception than pixel or patch methods. Furthermore, we propose a learnable MDGCN model to reason complex topological relations at different scales in superpixel-level space and enhance the contrast of object semantics by aggregating local difference information between central nodes and adjacent nodes. Modeling at the superpixel-level causes graph nodes to focus more on object regions while ignoring individual pixel features in the local space. As a result, we employ a bottom-up learning approach to enrich multi-level semantic representations of vision by integrating complementary spatial information from GCNs and CNNs. Experimental results validate the effectiveness of our proposed model on three downstream subtasks and their public benchmark datasets, respectively. The findings of this research raise two points for further consideration. One is that the MDGCN by dynamic graph mechanisms to improve the learning of nodes in different layers, and the other is to further explore advanced feature fusion strategies.

\vspace{-25pt}
\vfill
\end{document}